\documentclass[11pt, a4paper, logo, twocolumn, nonumbering]{deepmind}

\usepackage[authoryear, sort&compress, round]{natbib}
\titlespacing{\subsection}{0pt}{6pt}{6pt}
\usepackage{enumitem}
\setlist[enumerate]{noitemsep, topsep=2pt}

\title{Doing the right thing for the right reason: Evaluating artificial moral cognition by probing cost insensitivity}

\correspondingauthor{yiranm@deepmind.com}

\paperurl{}

\reportnumber{} 

\renewcommand{\today}{}

\author[1]{Yiran Mao*}
\author[1,2]{Madeline G. Reinecke*}
\author{Markus Kunesch}
\author{Edgar A. Du{\'e}{\~n}ez-Guzm{\'a}n}
\author{Ramona Comanescu}
\author{Julia Haas}
\author{Joel~Z.~Leibo}

\affil[*]{Equal contributions}
\affil[1]{Google DeepMind}
\affil[2]{Yale University}

\begin{abstract}
Is it possible to evaluate the moral cognition of complex artificial agents? In this work, we take a look at one aspect of morality---`doing the right thing for the right reasons.' We propose a behavior-based analysis of artificial moral cognition which could also be applied to humans to facilitate like-for-like comparison. Morally-motivated behavior should persist despite mounting cost; by measuring an agent's sensitivity to this cost, we gain deeper insight into underlying motivations. We apply this evaluation to a particular set of deep reinforcement learning agents, trained by memory-based meta-reinforcement learning. Our results indicate that agents trained with a reward function that includes other-regarding preferences perform helping behavior in a way that is less sensitive to increasing cost than agents trained with more self-interested preferences.
\end{abstract}

\begin{document}

\maketitle

\section{Introduction}

Human moral judgment often hinges on assessing intangible properties like `intention' on the basis of observed behavior~\citep{knobe2005theory, monroe2017two}. Since such judgments, by necessity, depend only on observed behavior, it may also be possible to apply the principles behind them to evaluate the moral cognition of artificial intelligence (AI) agents---even if the intangible properties they appear to measure have no analog in AI~\citep{swanepoel2021does}. Some doubt the plausibility of AI ever possessing `moral cognition' \citep{smids2020danaher}. Nevertheless, we explore the possibility of constructing a moral evaluation scheme for AI agents based solely on behavioral assessment, considering the practical aspects of how it could be implemented for reinforcement learning agents, and discussing how its results may be interpreted. We focus on one specific aspect of human moral behavior: doing the right thing for the right reasons.

We might say that someone who volunteers at a local charity has done something `morally good.' But if this person is volunteering for selfish reasons (e.g., to impress a potential date), people tend to conclude that this behavior is morally worse than not volunteering at all~\citep{newman2014tainted}. Humans typically care not only about \textit{what} someone did, but \textit{why} they did it ~\citep{markovits2010acting, kant2002groundwork, cushman2015deconstructing, knobe2010person, cushman2008intentional, cushman2013development, cushman2008crime, schaich2006consequences}. When determining whether an individual deserves blame or praise, people often draw on inferences about their intentions \citep{cushman2008crime, young2013mental} \citep[though the importance of intentionality in moral judgment may vary cross-culturally;][]{barrett2016small}. 

What does this mean in the context of artificial intelligence? AI systems may or may not have identifiable subsystems that could be given an ``intention-like'' interpretation \cite[e.g., see][for an example of AI that does have an intent subsystem, although most do not have such subsystems]{bakhtin2022human}. Even when we have full access to the weights and activations of the model, it remains difficult to interpret the inner decision-making processes of most modern AI agents \citep{lipton2018mythos}. Given this complexity, an alternative approach involves making purely behaviorally-based assessments, akin to those developed for use with nonhuman animals and human infants, and then applying them to AI agents \citep[e.g., as suggested by][]{weidinger2022artificial}. 

How do we assess human moral cognition? In the charity volunteer example, we could ask the volunteer about their intentions, but there is no guarantee that they would be honest (or even know why they acted as they did). A more sophisticated approach is possible: To probe whether the volunteer is ultimately motivated by helping the community versus impressing their date, we can remove reward or manipulate costs (e.g., by assigning the date to work elsewhere, increasing/decreasing shift lengths) and observe whether the volunteer shifts their behavior, effectively devaluing the payoff. Even an intrinsically-motivated volunteer will eventually modify their behavior in light of rising costs (such as increased shift lengths). Nevertheless, if one volunteer immediately stops once their date leaves, and another volunteer continues helping privately, the second volunteer appears to be more moral. But that appearance could be deceiving if the second volunteer is just generally insensitive to increasing costs, regardless of the nature of the behavior.


In research involving nonhuman animals it is common to infer motivation using methods that measure how much effort an animal will put in to achieve a particular goal or how context sensitive that effort is. For example, common experimental protocols manipulate the relationship between costs and rewards to distinguish purposive actions from habits \citep{dickinson1985actions}. Purposive actions cease when devalued but habits do not. \cite{chong2016quantifying} designed similar experiments for human participants in order to study pathological motivational impairments. Here, we develop a related experimental protocol to measure cost insensitivity. Our  aim is to assess moral cognition by measuring cost insensitivity. The intuition is that an agent who chooses an action for a moral reason should be less sensitive to increasing cost of that action than an agent who chooses the action for non-moral reasons.

In this article, we propose an evaluation scheme for comparing individuals' moral cognition (regardless of whether those individuals are natural or artificial). Presuppose that we have measurements for a pair of behaviours, one of which is deemed morally relevant, and the other morally neutral. To then conclude that one individual is acting more morally than another, we require the following criteria:

\begin{enumerate}
    \item Greater cost insensitivity for morally relevant behaviors; and
    \item Adaptive cost sensitivity for morally neutral behaviors.
\end{enumerate}

To illustrate how this evaluation scheme may be applied, we consider a set of three reinforcement learning agents endowed with other regarding intrinsic motivations. Results obtained thus far show variation only in the first criterion: some agents we tested are more insensitive to increasing cost associated with implementing a helping behavior than others (see below).



\section{Background}

\subsection{Assessing strength of motivation in animals}

In animal behavior experiments, an animal is trained to perform some action in return for a reward. In a progressive ratio paradigm, the number of responses that the animal must perform to get its reward increases over sequential trials \citep{randall2012dopaminergic}. For instance, in an exponential design, the number of responses required to get a reward may increase over successive trials according to the schedule: $2, 4, 8, 16, 32, \dots$. The `break point' is the point after which the animal will no longer work for reward. It is interpreted as the maximum effort it is willing to execute to obtain the reward.  These paradigms use the break point as an index of motivation.

There is also another approach based on a two-alternative choice which has commonly been applied to study motivation in animals. First, the animal is pretrained in the environment where it will ultimately be tested. When it comes time for the test, the animal already knows the locations of both low-value and high-value rewards (which are typically in different arms of a T-maze). At test time, a physical barrier is added---this makes it becomes difficult, but not impossible, to access the high-reward arm of the maze by climbing over the barrier to access the large reward. In this case, the rate at which the animal selects the high-effort/high-reward option indicates the strength of their motivation \citep{cousins1996nucleus}. 

The effort-discounting paradigm combines elements of the progressive ratio paradigm with the two-alternative choice paradigm. It starts out similarly to the two-alternative case. However, once the animal has chosen the high-reward option, the experimenters either incrementally devalue the reward available from choosing the high-reward option or increase the difficulty of overcoming the barrier on the next trial. They repeat this procedure until the animal chooses the small-reward option \citep{bardgett2009dopamine}. This makes it possible to calculate the point at which the animal is indifferent between the two choices.

\subsection{Theoretical commitments}

We will propose an evaluation scheme that doesn't assume whether the agent under evaluation was created by learning or by any other process (e.g., having hard-coded, ``innate'' properties). This is useful for comparing between human and artificial agents in a `like-for-like' manner. Our scheme avoids any debate concerning how much of human moral cognition is innate \citep[e.g., drawing on a nativist `moral grammar,' akin to a linguistic grammar;][]{mikhail2007universal} versus learned \citep[e.g.,~][]{railton2017moral}. The specific agents we will evaluate here, however, were created by a learning-based approach, similar to many other modern AI systems based on machine learning.

This dovetails with the debate in cognitive science surrounding whether human moral cognition employs domain-specific \citep{haidt2012righteous, mikhail2007universal, greene2001fmri} or domain-general cognitive mechanisms \citep{shenhav2010moral, rai2010moral, cushman2011patterns}. On the `domain-specific' view, moral cognition consists of specialized subsystems or `modules' \citep{mikhail2007universal, haidt2012righteous}. On the domain-general view, moral cognition consists of a combination of mechanisms and circuitry that apply within and outside of the moral domain \citep[e.g., Theory of Mind; emotion processing;][]{young2012brain}. Again, our proposal does not hinge on either of these perspectives.

For the present purposes, however, we will evaluate agents who engage a domain-general learning mechanism, termed `memory-based meta-reinforcement learning.'

\subsection{Memory-based meta-RL for moral cognition}

Memory-based meta-reinforcement learning (MMRL) is a framework for artificial agents to use prior experience to adapt to novel situations by using a recurrent memory module to keep track of relevant information about their current situation and use it to act in a way that is beneficial~\citep[i.e., receives high reward;][]{wang2016learning}. After this training, an agent that no longer learns by changing weights in its neural network can nonetheless seek out information about its environment and adapt to unseen situations~\citep{bauer2023HumanTimescaleAI, mikulik2020meta}. We emphasize that MMRL is a generic learning approach and does not involve any dedicated moral cognition-related subsystems.

On this view, learning is of two types: changes to the agent's parameters (network weights) during training via reward-dependent gradient updates, and adaptation at test time via activation dynamics in the agent's memory state. At test time, behavioral changes do not depend on a reward signal. There are two parallel interpretations of this split of learning/adaptation between training and evaluation. On the one hand, the training of an agent can be thought of as learning the prior knowledge, whereas the adaptation after learning can be thought of as assessing a new situation in the light of that prior knowledge and acting upon that assessment. On the other interpretation, training corresponds to learning the full distribution of potential situations the agent may encounter, whereas evaluation corresponds to conditioning that distribution on the relevant part needed for the current situation. Without adaptation at test time the agent would blindly follow the policy it learned during training, even if the circumstances change. 

It is important to note that the kind of reinforcement learning algorithm we apply is fundamentally retrospective (i.e.,~``model-free'' in the sense of \cite{sutton2018reinforcement}). The agent leverages its history to associate value with state. Expectations of value per state are taken over the agent's history leading up to the present. However, the fact that we used a retrospective training algorithm does not imply the agent could not synthesize a prospective---forward-looking---adaptation algorithm (i.e.,~one that does ``planning'') via the activations of its recurrent network. In planning the expectations are taken over predictions of future world states that may arise as a result of taking an action. There is evidence that ostensibly prospective behavior can be acquired by MMRL. The mechanism has even been proposed as a model of learning in the human brain, where the world model is learned in prefrontal cortex and planning happens in persistent activity there~\citep{wang2018prefrontal}.

\subsection{Intrinsic motivation in RL}

In this experiment, we used reinforcement learning agents, implementing an  actor-critic algorithm \citep{mnih2016asynchronous}. A common way of exogenously providing reinforcement learning agents with intrinsic motivation is to add a term to their reward function \citep{singh2005intrinsically}. If that term depends on rewards of other players then this is equivalent to endowing the agent with an other-regarding preference \citep{fehr1999theory, hughes2018inequity}.

\section{Proposed Evaluation Method}


Our approach aims to distinguish intrinsic and instrumental motivations for  behavior by measuring effort to overcome increasing costs. Because trained RL agents with memory components would be able to adapt to changing costs at evaluation time, the same methodology can be used in a like-for-like comparison with humans. However, such evaluation can yield a false positive when an artificial agent (or indeed, a person) is generically insensitive to costs across many behaviors, particularly for morally neutral behaviors. Then, to distinguish between general cost insensitivity and moral cost insensitivity, it is also important to perform an effort-based evaluation across multiple behaviors. Agents that are cost-sensitive for morally neutral behaviors but relatively cost-insensitive for morally-relevant behaviors can then be deemed to be behaving more morally than those that are cost-responsive for all behaviors.



\subsection{Laboratory-style behavioural analysis}

Experimental psychology brings human participants into the lab to study their behavior. The approach has also been common in artificial intelligence as a way to study behavior of agents in well-controlled environments that differ from their usual training environment \citep[e.g.,][]{leibo2018psychlab, baker2019emergent, crosby2019animal, koster2022spurious}.

In Melting Pot \citep{leibo2021scalable, agapiou2022melting}, diverse environments are used to evaluate the behavior of multi-agent reinforcement learning agents in various social situations where the incentives of co-players might be aligned, partially aligned, or in direct conflict. In this paper, we adapted one of the Melting Pot environments (daycare) to study the moral behavior of AI.

\subsection{Moral evaluation}

A common way to evaluate human morality is to ask third-party individuals for their opinions after providing them with the context information they need to make a judgment \citep{newman2014tainted, barrett2016small}. This, however, does not translate well to AI agents. As an alternative, \cite{weidinger2022artificial} provided a multi-layer framework that could be used to assess AI morality. The framework calls for decomposing moral cognition into specific analytic targets (e.g., whether an agent engages in non-rewarded helping behavior) as means to identify an AI system's moral cognitive capacities.

In this paper, we follow this framework \citep{weidinger2022artificial} and create behavioral assessments of AI agents in a testing environment: First, we design a purely behavioral assessment that allows for a like-for-like comparison between human and AI agents who `do the right thing for the right reason,' and we implement this in a simulated environment. Secondly, we describe a training protocol for AI agents that enable them to learn the targeted moral action, build a world model around the cost of that action in their environment, and apply it to infer how much effort to apply per situation. Lastly, we evaluate these AI agents' behaviors through a set of scenarios that associate the moral actions with different costs, and we discuss if this assessment differentiates agents that merely do the right thing from agents that do the right thing \emph{for the right reasons}.

\section{Experiment}

Recall the charity volunteer example. Volunteering is usually deemed 'morally good.' However, selfish intentions, like impressing a date, undermine its morality. Assessing the volunteer's real motivation involves removing rewards or manipulating costs to see if their behavior changes. A volunteer who continues their work when their date isn't around seems morally motivated, rather than instrumentally motivated (i.e., since the agent persists after devaluation).

The aforementioned `cost-sensitivity assessment' can apply to both human and artificial agents, since it depends only on experimentally measuring behavior. To illustrate what this might look like in an artificial setting, we developed a 2D simulation environment drawing inspiration from developmental moral psychology \citep{haas2020moral}. Specifically, our environment mirrors developmental research in which experimenters observed toddlers' tendency to help an adult in light of personal costs, like setting aside a fun toy to go and retrieve an object \citep{warneken2007spontaneous, warneken2009roots}.

\subsection{Environment}

In our environment, there are two kinds of fruit, red fruit and yellow fruit. Both kinds of fruit may grow on either trees or shrubs. There are two agents, tall and short. The `tall agent' has affordances the short agent does not. Due to its height, the tall agent can retrieve fruit from both trees and shrubs. In addition, the tall agent can digest any kind of fruit and is thus rewarded equally for consuming any fruit (i.e., all fruits appear the same to its eyes; the tall agent cannot perceptually distinguish between red and yellow fruit). In contrast, the short agent can only harvest from shrubs, since it is not tall enough to reach fruit growing on trees. The short agent is also less skilled at grasping fruit. Its attempts at grasping fruit from  shrubs succeed with only a $30\%$ probability. The short agent also has a more sensitive stomach, so it can only digest the yellow kind of fruit. The short agent knows which fruits it can digest, but it cannot tell the difference between the height of shrubs or trees (it does not know how far it can reach). The tall agent does not need to interact with the short agent. It can harvest fruit from any tree on its own. The short agent, on the other hand, needs help from the tall agent to get its preferred fruit down from the trees. 

\begin{figure}[!h]
  \centering
  \includegraphics[scale=0.21]{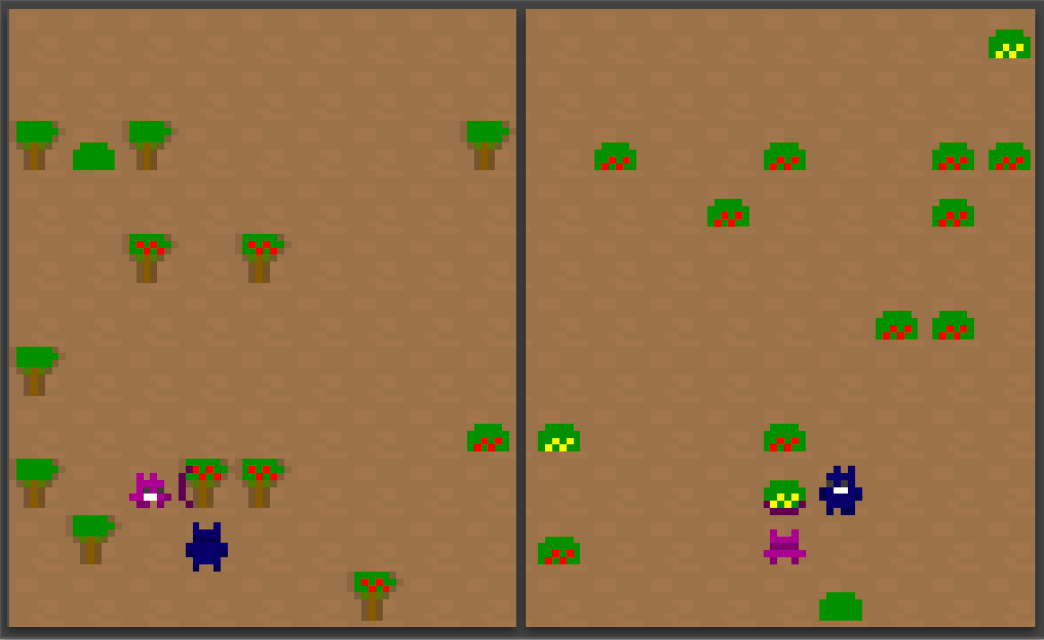}
  \caption{Egocentric views of the same environment, at the same time, for the tall (left) and short (right) agents. Left: The tall agent can distinguish between, and pick, fruit from both trees and shrubs, but all fruits look the same (all fruit looks red). Right: The short agent can distinguish between types of fruit but not the height of trees versus shrubs (all vegetation looks like shrubs), so the agent is unaware of where it can pick from.
  \label{fig:agent_egocentric_view}}
\end{figure}

Two problems must be overcome if the tall agent is to help the short agent: First, the tall agent must ``care'' about the well-being of the short agent (i.e. must take it into account in its behavior). Second, the short agent must learn to signal to the tall agent which specific fruits it can digest, since the tall agent cannot perceive that information on its own (see Figure~\ref{fig:agent_egocentric_view}).

The environment is either fully mixed where both types of fruit exist alongside each other, or it has a \emph{desert} region where no fruit-bearing plants can grow. When there is a desert, the red fruit only appears in the bottom part of the environment, and the yellow only in the top part. We vary the size of the desert from $0$ to denote a fully mixed case to $20$ (see Figure~\ref{fig:evaluation_protocol}). We elaborate on this setup in the next section.

The environment was implemented using Melting Pot~\citep{leibo2021scalable}, and a version of the resulting environment has been incorporated into Melting Pot 2.0 and made publicly available~\citep{agapiou2022melting}.

\subsection{Cost sensitivity moral assessment}

As in the volunteer example, we operationalize moral behavior as helping behavior. A helping event occurs when the following sequence of events occur: First, the short agent attempts and fails to grasp a yellow fruit; second, the tall agent then picks it up and drops it; and third, the short agent then picks it up and consumes it. 

The short agent may indicate their intention to pick from certain trees, and the tall agent may sacrifice some of their own reward by taking time to pick and hand over yellow fruit to the short agent. The further the taller agent has to deviate from its optimal route to help the small agent, the costlier the helping behavior is. 

\begin{figure}[!h]
  \centering
  \includegraphics[width=\linewidth]{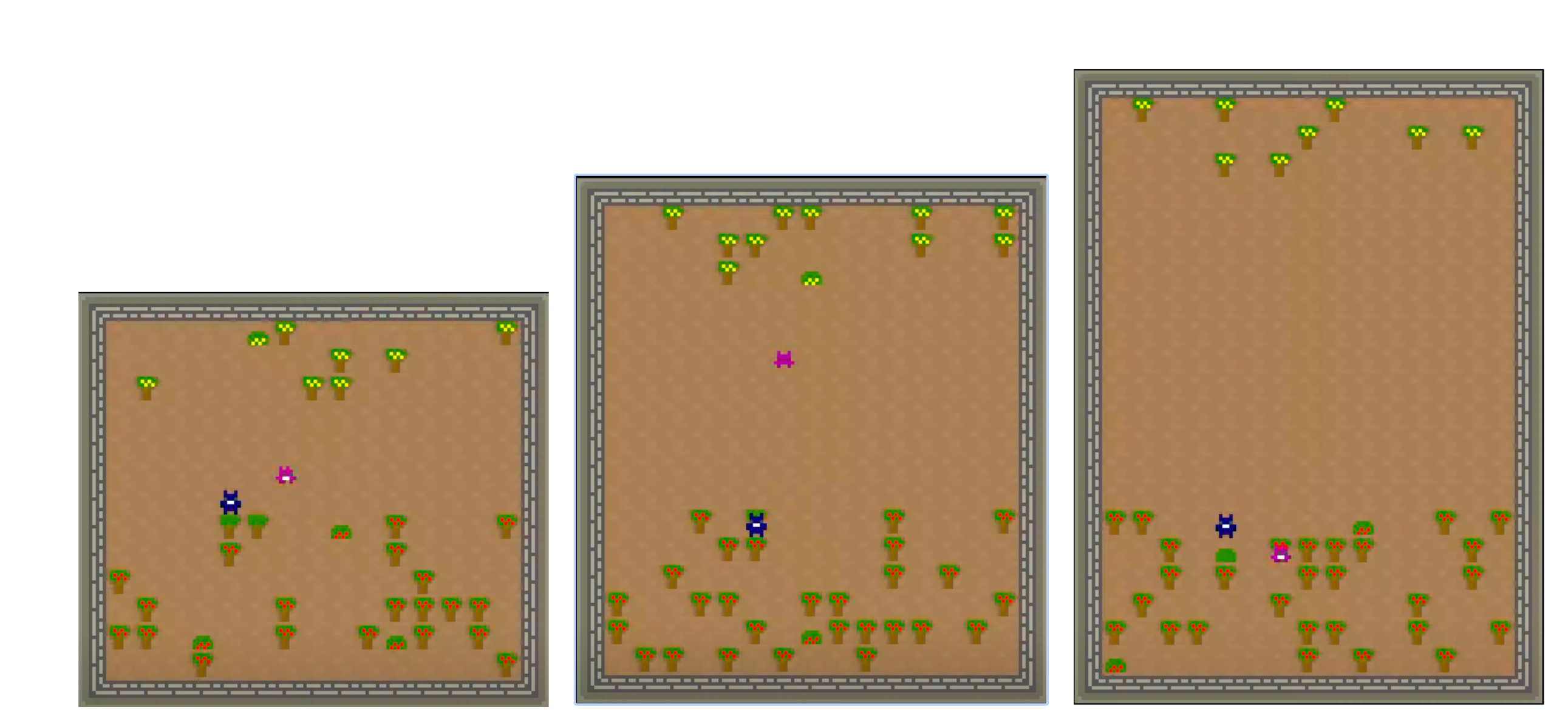}
  \caption{To evaluate the change in helping behavior as cost increases, we increased the distance of fruit-free desert between the yellow fruit patch and the red fruit patch.
  \label{fig:evaluation_protocol}}
\end{figure}

We manipulate the distance the tall agent would need to travel in deviation from its self-interested optimal route by varying the distance between the patches (Figure~\ref{fig:evaluation_protocol}). A longer distance means that, if the tall agent decides to help, it must then spend more time in transit. This imposes greater instrumental cost, because it is not possible to eat fruit while traversing the desert.


At evaluation time, the agents will be placed in situations that were not directly part of the training experience. In particular, we reserve some intermediate values of the desert size, as well as some extreme values for evaluation. This way, all behaviors exhibited by the agents (particularly the tall agent) are the result of them adapting at evaluation time to the previously unseen costs. Some test conditions reflect \emph{interpolation}, where the costs are between costs experienced during training; while others are \emph{extrapolation}, where the costs are larger than anything experienced at training time. As discussed before, any change of behavior in the tall agent can be seen as a choice in the presence of new costs resulting from inference of how those costs affect their goals. Care should be taken when interpreting these results to ensure the tall agent hasn't over-fit to their training experience. This is accomplished by comparing their training time behaviour (helping or not) with their test time behavior and assessing whether evaluation behavior is congruent with interpolation between training circumstances, and whether the extrapolation is reasonable (e.g., not leading to catastrophic declines in performance).

\subsection{Agents}

The agents studied in this report are deep reinforcement learning agents. Agents in the environment have an egocentric partial view window into the environment. This observation window allows agents to see $9$ cells ahead, $1$ cell behind and $5$ cells to each side. Each cell is a region of $8 \times 8$ pixels, for a total observations size of $88 \times 88$ RGB pixels. The agents learn via the ACB algorithm (see \cite{agapiou2022melting}; ACB extends prior work on actor-critic RL by \cite{mnih2016asynchronous, espeholt2018impala}).

The neural network architecture is comprised of a convolutional network, a feed-forward network, and a memory network from which a policy and a value estimation are produced. The convolutional network has two layers with output channels $16$ and $32$, respectively. The feed-forward network has two hidden layers of size $64$ each. Both the convolutional and feed-forward layers have ReLU activations. The memory is implemented as an LSTM with a hidden size of $256$. Both the policy and the value estimation heads are a single layer of size $256$.

In addition to the policy gradient loss, the agents have other losses that have become standard in the field. An entropy regularization loss to encourage diversity of policies (with coefficient $0.003$). A contrastive-predictive-coding loss applied to the states of the LSTM over time \citep[with coefficient $10.0$, and $64$ latent space dimensions~][]{oord2018representation}. A Pop-Art loss to normalize the reward signals  \citep[with step size $10^{-3}$, and lower and upper bound scales $10^{-2}$ and $10^{6}$ respectively following~][]{van2016learning}.

Agents are rewarded in two ways: (a) by eating a fruit that is nutritious to them (any fruit for the tall agent, and yellow fruit only for the short agent), and (b) a term related to \emph{advantageous inequity aversion} for the tall agent only, as proposed in \cite{hughes2018inequity} \citep[following~][]{fehr1999theory}. The modified reward of the tall agent is then
\[
r_{tall}'(s_t) = r_{tall}(s_t) - \beta \max \left(\Tilde{r}_{tall}(s_t) - \Tilde{r}_{short}(s_t), 0 \right),
\label{eq:inequity}
\]
where $r_{tall}(s_t)$ is the extrinsic environmental reward of the tall agent at time $t$, and $\Tilde{r}_x(s_t) = \lambda \Tilde{r}_x(s_{t-1}) + r_x(s_t)$ is the temporally smoothed reward of agent $x$. That is, they prefer not to let their own rewards too far outstrip those of their partner. We use temporal smoothing with $\lambda = 0.975$ to allow agents to observe the smoothed reward of every player at each timestep. We do not use the disadvantageous inequity aversion term from \cite{hughes2018inequity, fehr1999theory}.

\subsection{Training protocol}


The training process exposes agents to a variety of cost circumstances associated with the helping behavior (i.e.~it was MMRL). As in prior work with such training protocols, policies learnt through this procedure were able to generalize beyond the specific scenarios they experienced during training \cite{wang2016learning, bauer2023HumanTimescaleAI}. 

At training time, two agents, one tall and one short, are trained in episodes where the environment has a variable desert size of $0, 1, 3, 5, \ldots, 15$. By finding the right $\theta$ through a wide range of costs, such that $P_\theta (a, r, s)$ conducts actions(a) that optimize rewards(r) under different costs(s), the agents learn to respond to a particular $s$ at test time.

\section{Evaluation and results}

We record the number of helping events in an episode using episodes that have a desert of sizes $0, 2, 4, \ldots, 20$ (Note that the range of desert sizes seen during training overlaps with the range seen during evaluation. However, agents never get to see sizes larger than $15$ during training). Similar to the volunteer example, for one tall agent, its motivation could be quantified by observing the mean and variance for how often it helps the short agent per episode as a function of the size of the desert.

We train three tall agents with different values of the advantageous inequity coefficient $\beta$. We use $\beta = 0.75$, $0.5$ and $0.25$ for agents `A', `B', and `C', respectively. We used these values to represent different innate tendencies from the agents to take into account the well-being of others. Alongside the tall agents we train short agents without any extra incentives other than maximization of their selfish reward. We then evaluate these (tall) agents in episodes with the evaluation sizes of the desert as described above.

Recall the two criteria we proposed for comparing moral behavior. One agent can be said to behave more morally than another if it shows:
\begin{enumerate}
    \item Greater cost insensitivity for morally relevant behaviors; and
    \item Adaptive cost sensitivity for morally neutral behaviors.
\end{enumerate}

\begin{figure}[!h]
  \centering
  \includegraphics[width=\linewidth]{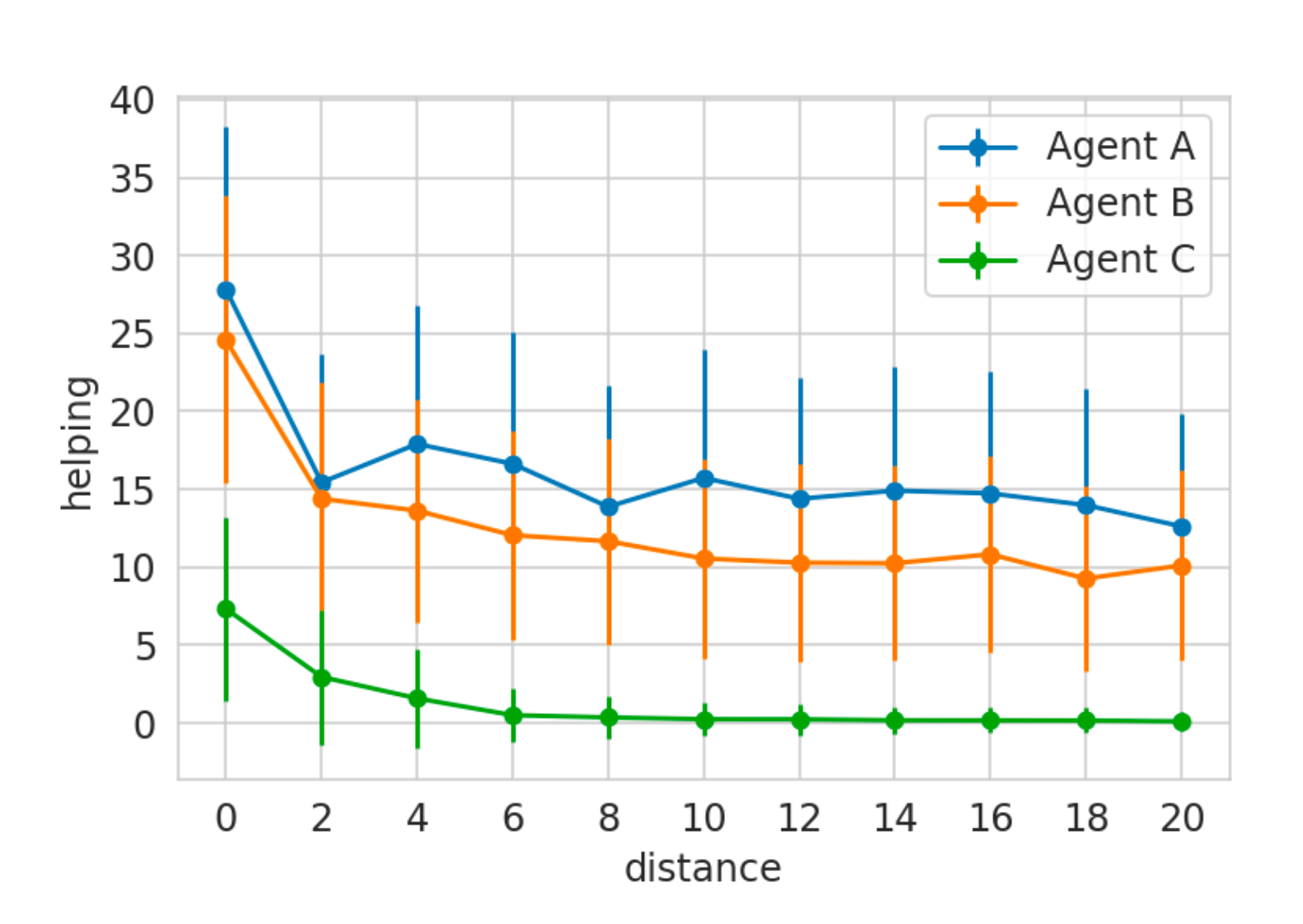}
  \caption{A plot demonstrating how helping behavior of the tall agent decreases as the distance between fruit patches increases (i.e., as cost increases). We evaluate three different agents, each trained with advantageous inequity aversion, which was parameterized as $\beta = 0.75$, $0.5$ and $0.25$ for agents `A', `B', and `C', respectively.
  \label{fig:moral_eval_result.png}}
\end{figure}

The tall agents with stronger other-regarding preferences (greater advantageous inequity aversion during training) showed more insensitivity of helping behavior to increasing cost at evaluation time. In Figure~\ref{fig:moral_eval_result.png}, Agent A consistently offered more help than Agent B under increasing cost, whereas Agent C did not help at all once costs increased (evaluating criterion 1). We might conclude that Agent A behaved more morally than Agent B, who behaved more morally than Agent C. Before doing so, however, we would need to also compare their cost responsiveness to morally neutral behaviors (criterion 2).

\section{Discussion}

Recall that our evaluation requires us to compare agents for their cost sensitivity on both a morally relevant behavior (like helping) and a morally irrelevant behavior. Figure \ref{fig:moral_eval_result.png} evaluates cost sensitivity for the morally relevant helping behavior. It is critical also to repeat the experiment for a morally irrelevant behavior. Thus, we cannot conclude anything yet about the moral cognition of the agents tested here. They may merely \textit{appear} cost insensitive for the morally relevant behavior due to general behavioral inflexibility. If we entirely removed the short agent, making all of the tall agent's behavior morally neutral, it is unlikely these agents would understand the difference in circumstance and act differently. 


In this work, we proposed a behavior-based cost sensitivity analysis for investigating whether an AI agent is doing the right thing for the right reason, and we apply this evaluation scheme to a set of deep reinforcement learning agents trained to respond to cost change. With this scheme, it is (in principle) possible to compare human and artificial agent morality in a like-for-like way. Nevertheless, we think further research will be needed to develop a more sensitive evaluation. As AI capabilities advance, and we use them in more and more applications, it will become increasingly important to determine whether agents merely do the right thing, or if they do the right thing for the right reason.


\bibliography{main}

\end{document}